\documentclass[10pt]{article}
\usepackage[margin=1in]{geometry}
\usepackage{iftex}
\ifPDFTeX
  \usepackage[T1]{fontenc}
  \usepackage[utf8]{inputenc}
  \usepackage{times}
\fi
\usepackage{natbib}
\usepackage{amsmath,amsfonts,bm}

\def\eqref#1{equation~\ref{#1}}
\def\1{\bm{1}}

\DeclareMathAlphabet{\mathsfit}{\encodingdefault}{\sfdefault}{m}{sl}
\SetMathAlphabet{\mathsfit}{bold}{\encodingdefault}{\sfdefault}{bx}{n}

\usepackage{amsmath,amssymb,booktabs,graphicx,float}
\usepackage{algorithm}
\usepackage{algpseudocode}
\usepackage{hyperref,url}
\hypersetup{
  hidelinks,
  pdftitle={MGPO: Mutant Group Policy Optimization for Structure-Diverse Protein Reinforcement Learning},
  pdfauthor={Han Zhang and Aoran Wang and Yulin Li and Zhenyu Yang and Mingxiao Li and Jinouwen Zhang and Chen Tang and Yizhou Wang and Shixiang Tang},
  pdfkeywords={protein structure generation, reinforcement learning, protein language model, credit assignment, structural diversity}
}

\title{MGPO: Mutant Group Policy Optimization for\\
Structure-Diverse Protein Reinforcement Learning}

\author{%
Han Zhang\textsuperscript{1}, Aoran Wang\textsuperscript{1}, Yulin Li\textsuperscript{1},
Zhenyu Yang\textsuperscript{1}, Mingxiao Li\textsuperscript{1}, Jinouwen Zhang\textsuperscript{1},\\
Chen Tang\textsuperscript{1,2}, Yizhou Wang\textsuperscript{1,2}, Shixiang Tang\textsuperscript{1,2}\\[0.6em]
\small\textsuperscript{1}Shanghai Artificial Intelligence Laboratory \quad
\textsuperscript{2}The Chinese University of Hong Kong
}
\date{}

\begin{document}
\maketitle

\begin{abstract}
Proteins can admit multiple structurally distinct solutions for the same amino-acid sequence, making broad structural exploration important for protein generative models. Mutation offers a natural way to expand policy exploration beyond ordinary autoregressive sampling, but we identify a previously under-characterized failure mode of mutation-augmented reinforcement learning: evolutionary exploration produces heterogeneous structural neighborhoods with different reward offsets, so a single global baseline can assign negative advantages to mutants that are improving within their own neighborhood. We introduce Mutant Group Policy Optimization (MGPO), which couples structured evolutionary exploration with neighborhood-aware credit assignment in the protein-structure-token space. MGPO generates structurally diverse mutant populations through reward-blind mutation and quality-diversity search, organizes them using reward-independent structural geometry, and computes each mutant's advantage relative to its full structural neighborhood before representative compression. This preserves learning signals for promising mutant families that would otherwise be suppressed by global population statistics. On the CAMEO22 benchmark, MGPO substantially improves mean lDDT-CA and TM-score. Diversity is evaluated in both structure token space and decoded 3D space; the measurements show that MGPO can improve the diversity of generated protein structures. A matched same-layout comparison, together with local-only ablations, shows that neighborhood-local credit assignment accounts for most of the improvement, while the evolutionary search provides the diverse structural neighborhoods on which this mechanism operates. These results establish MGPO as a mutation-aware policy-optimization framework for jointly improving structural quality and maintaining population-level exploration.
\end{abstract}

\section{Introduction}
\label{sec:introduction}
Protein folding---inferring a protein's three-dimensional structure from its
amino-acid sequence---is among the hardest problems in biology.  Despite rapid
progress in structure prediction~\citep{jumper2021alphafold,abramson2024alphafold3,lin2023esmfold,hayes2024esm3,SimpleFold},
autoregressive models still do not perform folding natively: even the strongest
general-purpose language models currently delegate structure prediction to
specialized external predictors rather than generating a structure end-to-end.
We therefore study whether an autoregressive model can generate structure
directly, using DPLM-2's discrete structure tokens as its output
vocabulary~\citep{wang2025dplm2}.  Optimizing such a generator with
reinforcement learning, however, surfaces a credit-assignment problem once
exploration is augmented by mutation: mutants of the same sequence occupy
separated structural neighborhoods with different reward offsets, so a candidate
that improves within its own neighborhood ($R_i>b_g$) can still fall below the
global mean ($R_i<b_{\rm global}$) and receive a negative global advantage,
suppressing entire promising mutant families.

MGPO builds on two lines of work.  First, DPLM-2's structure tokenizer encodes
each residue as a $13$-bit binary vector via Lookup-Free Quantization
(LFQ)~\citep{yu2023magvitv2}, putting a structure token in one-to-one
correspondence with its $13$-bit code~\citep{wang2025dplm2}; we adopt this
representation so that an autoregressive model can emit structure tokens
directly.  On this code, bit-flip-style operations always yield a valid
structure token, but simple single-bit flipping alone does not diversify
structural exploration enough---which is why richer mutation and evolutionary
search are needed.  Second, PPO, GRPO, and TOPR provide the on-policy and
off-policy RL foundations and training framework for language
models~\citep{schulman2017ppo,shao2024deepseekmath,leroux2025topr}, which we
adopt.  The gap between these two lines is that their advantage baselines are
global: once mutation spreads samples across structural neighborhoods with
different reward offsets, a global baseline misreads a within-neighborhood
improvement as a global regression.  This is precisely the problem we address.

We introduce Mutant Group Policy Optimization (MGPO), to our knowledge the
first systematic application of evolutionary mutation to the RL fine-tuning of
protein structure generation.  MGPO implements nine mutation/evolutionary
search strategies---genetic algorithms, quality-diversity, DPP selection,
island models, MAP-Elites, and others---in the DPLM-2 structure-token space,
each supplying structurally diverse mutants under a matched evaluation budget.
To keep mutation-driven training stable, MGPO then groups these mutants by
structure rather than reward and scores each mutant against its own group mean,
so that a family improving within its own neighborhood is not suppressed by the
global baseline.  Our contributions are: (1) introducing mutation/evolutionary
search into protein-structure RL, with a systematic comparison of nine
strategies; and (2) identifying the local/global advantage sign-failure in
mutation-augmented training, together with structure-local credit assignment as
its remedy.  Overall, MGPO improves the exploration effectiveness, structural
diversity, and prediction quality of protein structure models, offering a new
RL solution to the most challenging folding task.

\section{Related Work}
\label{sec:related-work}
DPLM-2 represents protein backbone structure with discrete LFQ tokens and a
frozen coordinate encoder/decoder~\citep{wang2025dplm2}; MGPO uses this codec
as a mutation and evaluation interface rather than using DPLM-2 as the trainable
policy.  Group-relative RL methods such as GRPO~\citep{shao2024deepseekmath},
sequence-level clipping/reweighting methods such as TOPR and related off-policy
variants~\citep{leroux2025topr,zheng2025gspo,shen2026vespo}, and large-scale RL
systems~\citep{yu2025dapo} assume rollout or otherwise characterizable behavior
distributions more directly than the evolutionary-mutant setting considered
here.  Diversity-aware protein optimization has also been pursued through set
rewards or regularizers~\citep{wang2025proteinzero,ye2026sgrpo}.  MGPO differs
by changing the reference used for mutant credit assignment instead of making
diversity itself the primary reward.  Evolutionary backends such as novelty
search, quality-diversity optimization, DPP selection, island models, and
MAP-Elites~\citep{lehman2011novelty,pugh2016quality,kulesza2012dpp,whitley1999island,mouret2015mapelites}
are treated as modular proposal mechanisms and are evaluated separately in the
appendix.

\section{Preliminaries}
\label{sec:background}
\subsection{Structure-token representation}
\label{sec:dplm2-tokenizer}
\label{sec:lfq-bits}
\label{sec:why-token-mutation}
A prompt $x$ fixes a single-chain amino-acid sequence of length $L$.  The
trainable autoregressive policy emits one DPLM-2 structure token per residue,
$\mathbf t(y)=(t_1,\ldots,t_L)$ with $t_i\in\{0,\ldots,8191\}$.  DPLM-2 is
frozen and used only to encode reference structures and decode generated token
paths.  Because $8192=2^{13}$, every token corresponds to a 13-bit LFQ code.
We use normalized bit-Hamming distance
\begin{equation}
 d(\mathbf t,\mathbf t')=\frac{1}{13L}\sum_{i=1}^{L}
 \operatorname{popcount}(t_i\oplus t'_i)
 \label{eq:hamming}
\end{equation}
for reward-blind mutation, deduplication, and grouping.  This is a computational
geometry in token space, not a physical distance; all candidates that affect
training are decoded and scored structurally.

The policy entering RL is initialized from Qwen3.5-4B, whose vocabulary is
expanded with the 8192 structure tokens plus special tokens marking the start
and end of a protein structure, and is then trained as the RL cold-start model
on 200B tokens from the AlphaFold Protein Structure Database v3~\citep{varadi2022alphafold}.
We therefore call this model the
\textbf{SFT initialization} throughout and measure every RL delta relative to
it.

\subsection{Mutation-induced baseline mismatch}
\label{sec:setup}
For a mutant pool $\mathcal M$ partitioned by reward-blind structure into
$G_1,\ldots,G_{n_G}$, let $b_g=|G_g|^{-1}\sum_{i\in G_g}R_i$.  A global
baseline can reverse local improvement whenever
\[
 R_i-b_g>0 \quad\text{and}\quad R_i-b_{\rm global}<0.
\]
This mismatch is specific to heterogeneous pooled data; it does not imply that
global-baseline on-policy RL generally collapses.  Our controls therefore
separate the effect of mutation from the effect of the baseline.

\paragraph{Direct sign-reversal diagnostics.}
For each mutant we define the directional failure indicator
\begin{equation}
 \chi_i^{+\rightarrow-}=\mathbb I[R_i-b_{g(i)}>0\ \wedge\ R_i-b_{\rm global}<0],
 \label{eq:signflip}
\end{equation}
and report both its frequency and its local-advantage mass,
\begin{equation}
 \mathrm{FlipRate}=\frac{1}{|\mathcal M|}\sum_{i\in\mathcal M}\chi_i^{+\rightarrow-},\qquad
 \mathrm{FlipMass}=\frac{\sum_i |R_i-b_{g(i)}|\chi_i^{+\rightarrow-}}
 {\sum_i |R_i-b_{g(i)}|},
 \label{eq:flipmass}
\end{equation}
We also report the opposite-direction rate
$\mathbb I[R_i-b_{g(i)}<0\wedge R_i-b_{\rm global}>0]$ so that the audit does
not presuppose an asymmetric outcome.  These quantities are descriptive
mechanism diagnostics; the fixed-batch intervention in
Section~\ref{sec:exp-mechanism} tests whether the sign changes alter the actual
update direction.

\section{Mutant Group Policy Optimization}
\label{sec:methods}
\subsection{Grouping and structure-local advantages}
\label{sec:mgpo}
For each prompt, the rollout policy produces $n_A=8$ anchors.  The default
Batch-Cluster QD backend runs seven QD generations, each contributing $E=8$
structurally evaluated mutants, for a total of $n_M=56$ structurally evaluated
mutants.  Balanced reward-blind $k$-medoids partitions the mutants into
$n_G=3$ groups, and $n_R=8$ representatives per group are retained, yielding
$8+24=32$ training sequences.
Crucially, the baseline is computed on the \emph{full}
cluster before representative selection:
\begin{equation}
 b_g=\frac{1}{|G_g|}\sum_{j\in G_g}R_j,\qquad
 A_i^M=\frac{R_i-b_g}{s_M},\quad
 s_M=\operatorname{Std}_{j\in\mathcal M}(R_j).
 \label{eq:mutant-advantage}
\end{equation}
Using the full cluster prevents the reward-stratified selector from moving the
reference.  A shared $s_M$ keeps advantage magnitudes comparable across groups.
Figure~\ref{fig:mgpo-framework} summarizes the data flow; selection details are
in Appendix~\ref{app:mgpo-details}.

\paragraph{Exact reward accounting.}
In the default MGPO learning rule, every member of the $n_M=56$
mutant pool that enters Eq.~\eqref{eq:mutant-advantage} receives the
same \emph{full} structural reward $R$ used for training.
Cheap token-space or proxy scores may be used to
construct or prune an \emph{unevaluated} proposal pool, but they are not
substituted for $R_i$ when computing $b_g$, $s_M$, representative quality, or
mutant advantages.
An approximate survivor-only full-reward mode is evaluated separately
in Appendix~\ref{app:parallel-speedup}.

\begin{figure}[t]
  \centering
  \includegraphics[width=\linewidth,trim=0 0 44bp 0,clip]{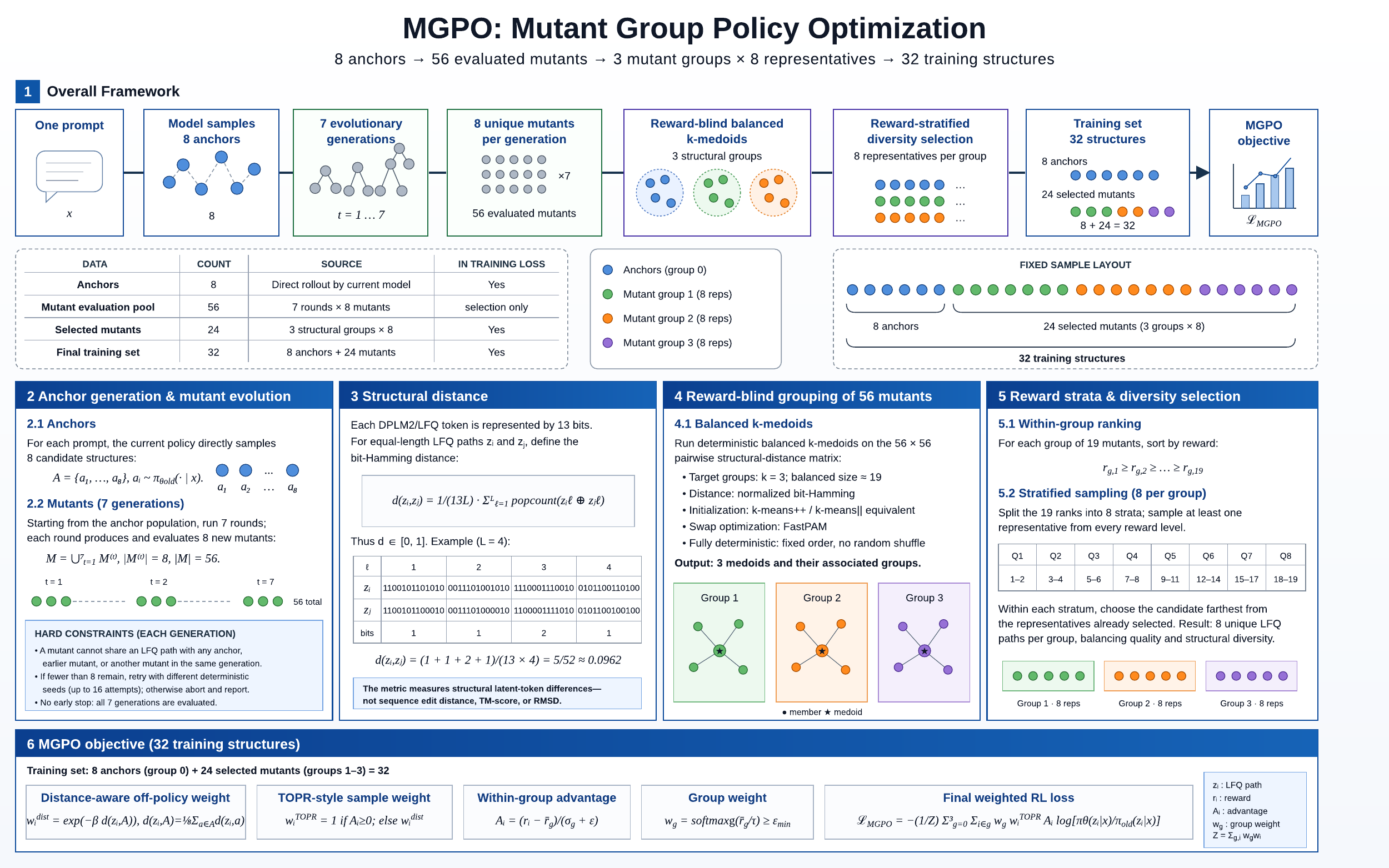}
  \caption{MGPO Framework. Eight rollout anchors seed 56 evaluated mutants;
  reward-blind structural grouping is followed by representative selection and
  structure-local mutant credit assignment. The key causal control keeps the
  entire layout fixed and replaces only $b_g$ with $b_{\rm global}$.}
  \label{fig:mgpo-framework}
\end{figure}

\subsection{Secondary weighting and distance stabilization}
The full implementation allocates every retained group nonzero learning mass,
\begin{equation}
 W_g=w_{\min}+(1-w_{\min})(n_G+1)p_g,
 \label{eq:group-weight}
\end{equation}
where $p_g$ is a numerically stable softmax over group means (Appendix
\ref{app:baseline}).  The distance heuristic uses the mutant's LFQ shift
$\delta_i$ from its root/current-policy anchors to construct
$q_i=\exp(-\kappa\delta_i)$.  Only negative mutant advantages are attenuated:
\begin{equation}
 \omega_i=\begin{cases}
 1,&A_i^M\ge0,\\
 q_i^{a}\exp\{b(1-q_i)\},&A_i^M<0.
 \end{cases}
 \label{eq:mutant-offpolicy-weight}
\end{equation}
We use $a=b=1$.  More generally, for $0<b\le a$ and $q\in(0,1]$,
$\log\omega^- =a\log q+b(1-q)\le(b-a)(1-q)\le0$, so the rule can only shrink,
not amplify, a negative update.  This does \emph{not} remove off-policy bias:
$q_i$ is a structural-shift score rather than a probability ratio.  We therefore
treat this component as an auxiliary stabilization heuristic rather than as a
theoretical off-policy correction.

\subsection{Training objective}
Anchors use the rollout-based TOPR sequence weight~\citep{leroux2025topr};
mutants use the local advantage, group weight, and optional attenuation above.
With $\bar\ell_\theta(y\mid x)=T_y^{-1}\sum_t\log\pi_\theta(y_t\mid x,y_{<t})$,
\[
 \ell_a=-W_0\omega_a^{\rm TOPR}A_a\bar\ell_\theta(a\mid x),\qquad
 \ell_i=-W_{g(i)}\omega_iA_i^M\bar\ell_\theta(i\mid x),
\]
\begin{equation}
 \mathcal L_{\rm MGPO}=\frac{\sum_{a\in\mathcal A}\ell_a+
 \sum_{g=1}^{n_G}\sum_{i\in S_g}\ell_i}{n_A+n_Gn_R}
 +\beta\mathcal L_{\rm KL}^{\rm anchor}.
 \label{eq:mgpo-loss}
\end{equation}
Mutant likelihoods are computed from the mutant paths themselves rather
than from the anchor paths.

\subsection{Evolutionary backend strategies}
\label{sec:backends}
The evolutionary backend is a modular search component: it supplies the 56
evaluated mutants and is otherwise independent of the credit-assignment
mechanism.  We implement nine mutation and survivor strategies that differ only
in their proposal and survivor mechanism; the downstream grouping, weighting,
and loss of Section~\ref{sec:mgpo} are identical.  Batch-Cluster QD is the
default.  All nine are matched in \emph{expensive structural evaluations} (not
necessarily in total CPU or proposal compute), so the comparisons below
attribute differences to the search mechanism rather than to a larger structure
budget.  Table~\ref{tab:nine-strategies} summarizes the nine strategies and the
single factor each one varies; full specifications are in
Appendix~\ref{app:backends}.

\begin{table}[t]
\caption{Evolutionary backend strategies.  Batch-Cluster QD is the default;
the downstream MGPO grouping and loss are unchanged.}
\label{tab:nine-strategies}
\centering\scriptsize
\setlength{\tabcolsep}{3pt}
\begin{tabular}{p{0.14\linewidth}p{0.20\linewidth}p{0.60\linewidth}}
\toprule
Backend & Varied factor & Proposal and survivor mechanism \\
\midrule
Standard GA & Fitness selection & CRF, LFQ, and random proposals; global elitist survivor selection. \\
Sequential QD & Local competition & Nearest-resident replacement plus a bounded novelty archive. \\
Batch-Cluster QD & Proposal compression & A cheap pool is reduced by $k$-medoids; survivors combine quality and max--min distance slots. \\
CAMP & Mutation location & Separated disagreement centers receive paired local and broad perturbations. \\
QDPP & Set diversity & Quality-weighted DPP selection for cheap proposals and survivors. \\
ISLE & Population topology & Four lineage islands use neighbor crossover and retain quality/diversity specialists. \\
AOPS & Operator schedule & Mutation operators are selected by UCB from past quality--novelty utility. \\
ELEX & Many-case selection & $\epsilon$-lexicase filters reward components and novelty cases. \\
MAPX & Structured archive & MAP-Elites cells index distance from the initial best and bit-density shift. \\
\bottomrule
\end{tabular}
\end{table}

\section{Experiments}
\label{sec:experiments}
\subsection{Setup, metrics, and causal controls}
\label{sec:exp-data}
The native-reference test set is the CAMEO22 benchmark~\citep{cameo22}, a
held-out set of single-chain targets.  RL uses the SFT initialization above.
End-to-end evaluation samples $K=8$
completions directly from the final policy; no mutation or reward-based
selection is used at test time.

\paragraph{Metrics.}
We report two quality metrics judged against the native backbone---lDDT-CA and
TM-score---and two population-diversity metrics.  The LFQ diversity is the mean
normalized bit-Hamming distance over all pairs of the $K=8$ sampled structures,
\begin{equation}
 \text{LFQ } \bar d=\frac{2}{K(K-1)}\sum_{1\le i<j\le K}d(\mathbf t_i,\mathbf t_j),
 \label{eq:bar-d}
\end{equation}
with $d$ defined in Eq.~\eqref{eq:hamming}; larger $\bar d$ means the sampled
structures are farther apart in token space.  The decoded 3D diversity is the
analogous mean pairwise $1-$TM-score,
\begin{equation}
 \text{3D div} =\frac{2}{K(K-1)}\sum_{1\le i<j\le K}\bigl(1-\mathrm{TM}(\mathbf t_i,\mathbf t_j)\bigr),
 \label{eq:d3d}
\end{equation}
where $\mathrm{TM}(\cdot,\cdot)$ is the TM-score between two decoded structures;
it is reported alongside $\bar d$ so that conclusions do not rely on LFQ
geometry alone.  Finally, collapse is defined
\emph{per target}: for target $j$, let $\bar d_j^{\rm init}$ be the LFQ
diversity of the SFT initialization and $\bar d_j^{\rm final}$ that of the
trained policy under matched sampling; target $j$ collapses if
$\bar d_j^{\rm final}<0.5\bar d_j^{\rm init}$, and the collapse rate is the mean
of this indicator over targets.

An online control we call \textbf{Same-layout global} matches full MGPO in the
number of anchors and evaluated mutants, reward-blind grouping algorithm,
representative-selection rule, group weighting, distance rule, and 32-sequence
training layout, but replaces every mutant $b_g$ by one global mutant baseline
$b_{\rm global}=|\mathcal M|^{-1}\sum_{i\in\mathcal M}R_i$.  Independent online
runs are initialized from the same SFT checkpoint and use matched training
schedules and seed sets where possible, but the realized pools are allowed to
diverge after the first different update.  The exact-sample causal comparison is
therefore performed separately by the fixed-batch intervention in
Section~\ref{sec:exp-mechanism}.

\subsection{Main quality--diversity results}
\label{sec:exp-main}
\begin{table}[t]
\caption{Policy-only endpoint results on CAMEO22.  ``Same-layout
 global'' matches the MGPO algorithmic layout but uses a global mutant baseline.
 All methods are evaluated from $K=8$ policy-only samples with the same frozen
 structural decoder and endpoint protocol.  ``Coll.'' denotes the collapse rate
 defined in Section~\ref{sec:exp-data}.}
\label{tab:main-results}
\centering\scriptsize
\setlength{\tabcolsep}{3pt}
\resizebox{\linewidth}{!}{%
\begin{tabular}{llrrrrr}
\toprule
Method & Training signal & lDDT avg@8 $\uparrow$ & TM avg@8 $\uparrow$ & LFQ $\bar d$ $\uparrow$ & 3D div. $\uparrow$ & Coll. $\downarrow$ \\
\midrule
SFT initialization & no RL & 0.7028 & 0.7061 & 0.0400 & 0.1280 & -- \\
GRPO-8 & 8 on-policy & 0.7246 & 0.7370 & 0.0388 & 0.1292 & 21.7\% \\
GRPO-64 & 64 on-policy & 0.7282 & 0.7410 & 0.0393 & 0.1310 & 20.4\% \\
\midrule
Mutant-GRPO & global, 8+56 pool & 0.7138 & 0.7264 & 0.0345 & 0.1180 & 27.8\% \\
Same-layout global & global, MGPO layout & 0.7145 & 0.7270 & 0.0352 & 0.1205 & 25.6\% \\
MGPO-local-only & local, no aux. weights & 0.7391 & 0.7496 & 0.0458 & 0.1452 & 10.9\% \\
MGPO (full) & local + aux. terms & \textbf{0.7427} & \textbf{0.7532} & \textbf{0.0486} & \textbf{0.1518} & \textbf{5.8}\% \\
\midrule
Mutant-GRPO+DivReward & explicit diversity reward & 0.7286 & 0.7417 & 0.0436 & 0.1380 & 17.2\% \\
ProteinZero-style & diversity regularizer & 0.7321 & 0.7446 & 0.0444 & 0.1412 & 14.8\% \\
SGRPO-style & set-diversity reward & 0.7360 & 0.7479 & 0.0461 & 0.1460 & 11.7\% \\
Mutant-AWR & reward-weighted imitation & 0.7334 & 0.7442 & 0.0418 & 0.1350 & 16.9\% \\
\bottomrule
\end{tabular}%
}
\end{table}
Table~\ref{tab:main-results} shows that structure-local credit assignment is the
main source of MGPO's gain.  Under the matched Same-layout global control,
replacing the global mutant baseline with structure-local credit improves both
quality metrics and both diversity metrics, while substantially reducing
collapse.  The local-only variant already recovers most of the improvement,
indicating that the local baseline is the dominant factor, with group weighting
and distance attenuation providing smaller secondary gains.  MGPO also
outperforms the on-policy, diversity-aware, and reward-weighted imitation
baselines, supporting the claim that its benefit comes from preserving useful
learning signals within heterogeneous structural neighborhoods rather than from
additional sampling or explicit diversity shaping.

\subsection{Mechanism audit: does the global baseline reverse useful updates?}
\label{sec:exp-mechanism}
The endpoint comparison in Table~\ref{tab:main-results} is complemented by a
mechanism audit on saved mutant pools.  At every logged training checkpoint we
compute the directional FlipRate and FlipMass from
Eqs.~\eqref{eq:signflip}--\eqref{eq:flipmass}, the opposite-direction flip rate,
and the fraction of selected representatives affected by each event.  We report
early/middle/late training thirds separately so that an endpoint average cannot
hide a transient failure mode.

\paragraph{Paired fixed-batch intervention.}
For a stratified set of checkpoints, we snapshot the same model parameters and
optimizer state and construct one shared batch containing the same anchors,
fully scored mutants, reward-blind groups, and representatives.  Two model
copies receive exactly one optimizer step: one with $b_g$ and one with
$b_{\rm global}$; every other loss term and random seed is fixed.  Let
$g_{\rm local}$ and $g_{\rm global}$ denote the mutant-branch gradients before
the optimizer transformation.  We report their cosine similarity, norm ratio,
and the mean post-step change in log probability of the
$+\!\rightarrow\!-$ sign-flipped mutants.  Because both copies use identical
model parameters, optimizer state, samples, grouping, representatives, loss
terms, and random-number state, the intervention isolates the effect of
replacing $b_g$ with $b_{\rm global}$ on the immediate update direction.

\begin{table}[t]
\caption{Mechanism audit from saved pools and fixed-batch interventions.
Values are averaged by target first and then summarized over early, middle, and
late training thirds.}
\label{tab:mechanism-audit}
\centering\small
\setlength{\tabcolsep}{5pt}
\begin{tabular}{lccc}
\toprule
Diagnostic & Early & Middle & Late \\
\midrule
$+\!\rightarrow\!-$ FlipRate (\%) & 18.6 & 22.4 & 25.7 \\
$+\!\rightarrow\!-$ FlipMass (\%) & 13.2 & 17.9 & 21.6 \\
$-\!\rightarrow\!+$ opposite FlipRate (\%) & 7.4 & 5.8 & 4.1 \\
Selected representatives flipped (\%) & 20.1 & 24.8 & 28.3 \\
$\cos(g_{\rm local},g_{\rm global})$ & 0.78 & 0.66 & 0.54 \\
$\|g_{\rm global}\|/\|g_{\rm local}\|$ & 0.94 & 0.89 & 0.84 \\
$\Delta\log\pi$ on $+\!\rightarrow\!-$ mutants: local step & +0.021 & +0.028 & +0.031 \\
$\Delta\log\pi$ on $+\!\rightarrow\!-$ mutants: global step & -0.014 & -0.022 & -0.027 \\
\bottomrule
\end{tabular}
\end{table}

Table~\ref{tab:mechanism-audit} shows that sign-flipped mutants are not a
negligible fraction of the pool, that their local-advantage mass grows over
training, and that the local and global mutant-branch gradients become
increasingly misaligned while the global step actively decreases the log
probability of the flipped mutants.

\subsection{Grouping granularity and mutation-budget sensitivity}
\label{sec:exp-group-sensitivity}
The central hypothesis requires structure-local neighborhoods, so the default
$n_G=3$ should not be treated as an unexamined constant.  We vary
$n_G\in\{1,2,3,4,6\}$ while holding the number of selected mutant training
sequences fixed by setting $n_R=24/n_G$, so every setting still uses 32 training
sequences including the anchors.
The $n_G=1$ condition is the natural
no-locality endpoint within this granularity sweep: it has one mutant group and
therefore one mutant baseline.  Its endpoint metrics numerically coincide with
the Same-layout global control in Table~\ref{tab:main-results}, as expected
when every mutant shares a single reward reference; the two settings differ in
how that reference is reached (single-group $k$-medoids here versus a global
pooled mean under $n_G=3$ grouping there) rather than in the resulting
baseline.

\begin{table}[t]
\caption{Pre-specified locality sensitivity.  The selected mutant count is held
fixed so the policy-update batch size is constant.}
\label{tab:group-sensitivity}
\centering\small
\setlength{\tabcolsep}{5pt}
\begin{tabular}{rrrrrrr}
\toprule
$n_G$ & $n_R$ & lDDT avg@8 $\uparrow$ & TM avg@8 $\uparrow$ & LFQ $\bar d$ $\uparrow$ & 3D div. $\uparrow$ & FlipRate \\
\midrule
1 & 24 & 0.7145 & 0.7270 & 0.0352 & 0.1205 & 0.0 \\
2 & 12 & 0.7348 & 0.7457 & 0.0397 & 0.1332 & 14.6 \\
3 & 8  & \textbf{0.7427} & \textbf{0.7532} & \textbf{0.0486} & \textbf{0.1518} & 22.4 \\
4 & 6  & 0.7419 & 0.7523 & 0.0405 & 0.1376 & 24.8 \\
6 & 4  & 0.7394 & 0.7502 & 0.0402 & 0.1366 & 27.1 \\
\bottomrule
\end{tabular}
\end{table}

Table~\ref{tab:group-sensitivity} shows that both quality and diversity peak
near the default $n_G=3$, while FlipRate increases monotonically with $n_G$.
The gain therefore tracks the amount of baseline heterogeneity rather than the
mere presence of clustering.

\subsection{Search-based imitation controls}
\label{sec:exp-imitation}
Because mutation produces samples without a tractable behavior density, we
include a reward-weighted self-training control to distinguish search-output
imitation from signed local credit assignment.  This control uses exactly the
same anchors, mutant pool, grouping, and selected mutant representatives but
removes signed local advantages.  For \textbf{Mutant-AWR}, selected mutants
receive normalized positive weights
\begin{equation}
 \alpha_i=\frac{\exp((R_i-\bar R_{\mathcal M})/\tau)}
 {\sum_{j\in S}\exp((R_j-\bar R_{\mathcal M})/\tau)},\qquad
 \mathcal L_{\rm AWR}=-\sum_{i\in S}\alpha_i\bar\ell_\theta(i\mid x),
 \label{eq:mutant-awr}
\end{equation}
with temperature $\tau$ selected once on validation and then frozen.  The
anchor branch, KL term, structural evaluation budget, and training layout are
unchanged.  Table~\ref{tab:main-results} shows that MGPO retains a clear
advantage over this control, so its benefit is harder to explain as merely
distilling high-reward search outputs.

\subsection{Geometry visualization and remaining audits}
\label{sec:exp-pca}
\begin{figure}[t]
  \centering
  \includegraphics[width=\linewidth]{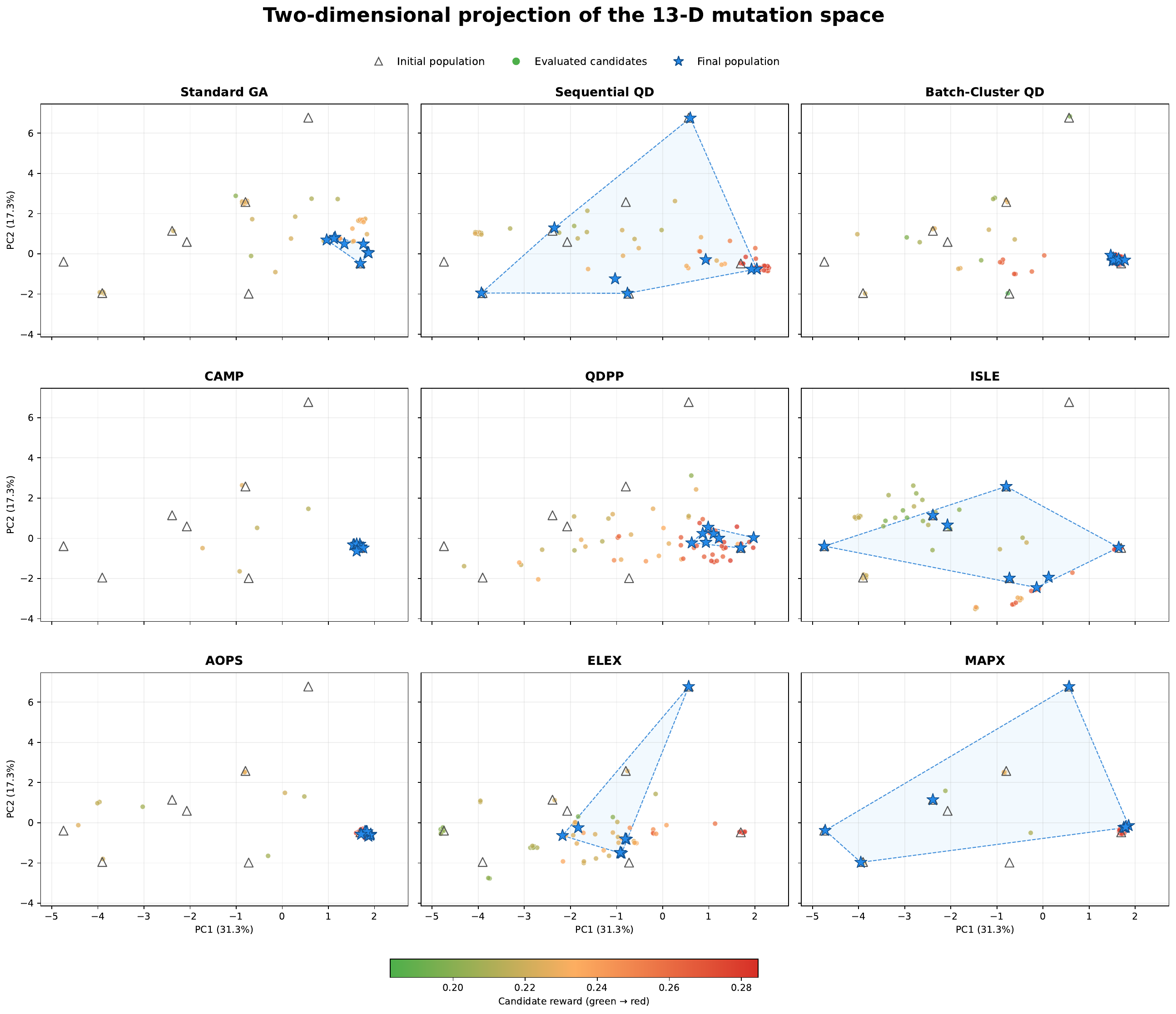}
  \caption{Shared-PCA view of a complete 64-evaluation trace.
  Features are per-position mean occupancies of the 13 latent token bits.
  Triangles denote the initial population, circles decoded candidates, and
  stars the final population; all nine panels use the same PCA basis and axis
  limits.  Candidate color encodes lDDT-CA.}
  \label{fig:pca}
\end{figure}

To inspect whether the diversity statistic corresponds to visibly different
search paths, we project one complete target trace (target 13) into a shared
two-dimensional PCA basis (Figure~\ref{fig:pca}).
The panel is illustrative; the quantitative conclusions rest on the
metrics in Tables~\ref{tab:main-results}--\ref{tab:mechanism-audit}.
Each backend contributes 8 initial anchors, 64
decoded candidate points, and 8 final survivors; each point is represented by
its 13-dimensional mean bit-occupancy vector (the average over residues of the
13 LFQ bits, a different feature space from the $13L$-bit flattened matrix used
for $r_{\mathrm{eff}}$).  The first two components explain $31.3\%$ and
$17.3\%$ of the pooled variance, respectively; the projection is therefore used
as a visualization and not as a replacement for the full Hamming metric.
Standard GA's final stars occupy a compact region, consistent with elitist
convergence.  Sequential QD and ISLE preserve broader regions, while QDPP
retains a high-reward final set without collapsing to a single cluster.

The remaining geometry, proxy, and backend audits are reported in
Appendix~\ref{app:geometry-audit}.
Appendix~\ref{app:geometry-audit} reports the within/between-group
decoded distances and the LFQ--3D adjusted Rand indices; those results
support interpreting reward-blind LFQ groups as algorithmic structural
neighborhoods rather than recovered biological states.
Appendix~\ref{app:backend-diag} reports the nine mutation-backend diagnostics,
including their quality--diversity trade-off.  Appendix~\ref{app:parallel-evolution}
reports the statistics and efficiency accounting, including the
correctness-preserving acceleration of the exact MGPO protocol and the
approximate proxy-reward engineering variant.

\paragraph{Statistical reporting.}
The statistical unit for endpoint evaluation is the held-out target.  Tables
\ref{tab:main-results} and \ref{tab:group-sensitivity} report aggregate means
over the same CAMEO22 benchmark under matched $K=8$ policy-only evaluation.  The
mechanism audit first aggregates diagnostics within target and then summarizes
them over early, middle, and late training.  The primary causal evidence comes from the large same-layout contrast
and the exactly matched fixed-batch intervention; smaller component-level differences are reported descriptively and
are not assigned statistical-significance claims.

\section{Conclusion}
\label{sec:conclusion}

We introduced Mutant Group Policy Optimization (MGPO), a mutation-aware policy-optimization framework that couples structured evolutionary exploration with neighborhood-aware credit assignment for protein structure generation. MGPO generates structurally diverse mutants through reward-blind mutation and quality-diversity search, groups them using reward-independent structural geometry, and computes each mutant's advantage within its complete structural neighborhood before representative compression. Because the baseline is computed on the full cluster prior to compression, MGPO preserves learning signals from mutants that improve within their local neighborhoods but would be suppressed by global population statistics. Same-layout comparisons and local ablations show that structure-local credit assignment is the primary source of improvement, while evolutionary search supplies the diverse structural neighborhoods on which this mechanism operates, and group weighting and distance terms serve as auxiliary mechanisms for stabilization and compression. Overall, this work makes two main contributions: it systematically introduces mutation and evolutionary search into reinforcement-learning fine-tuning for protein structure generation and compares nine mutation/evolutionary strategies; and it identifies the sign-failure between local and global advantages in mutation-augmented training and proposes structure-local credit assignment as its remedy. MGPO thereby improves the exploration effectiveness, structural diversity, and prediction quality of protein structure models, offering a mutation-aware policy-optimization framework for protein structure generation.

\section*{AI Use Statement}
Generative AI tools were used to assist with manuscript editing, scientific
figure preparation, code editing, and to provide feedback on experimental
design and interpretation of experimental results.  All AI-assisted
methodological suggestions, code modifications, and result interpretations were
reviewed and verified by the authors.  All reported experimental results were
produced by the authors' experimental pipeline, and the authors take full
responsibility for the content, claims, code, and experimental results in this
work.

\section*{Ethics Statement}
This work studies conditional protein-structure generation for sequences
provided as input and does not introduce a de novo amino-acid sequence-design
pipeline.  Generated structures are computational hypotheses rather than
experimental validation.  The method inherits the dual-use profile of protein
modeling, and any downstream practical use should apply the usual structural,
functional, and experimental verification procedures.

\section*{Reproducibility Statement}
The main text and appendices specify the MGPO objective, reward components,
evolutionary backends, sampling and structural-evaluation budgets, key training
hyperparameters, causal controls, and systems implementation used in the
reported experiments.

\bibliography{ds_mupo_references,ds_mupo_references_additions}
\bibliographystyle{iclr2027_conference}

\clearpage
\appendix

\section{Notation summary}
\label{sec:notation}
\begin{table}[H]
\caption{Notation used throughout the paper.}
\label{tab:notation}
\centering
\small
\begin{tabular}{ll}
\toprule
Symbol & Meaning \\
\midrule
$x$, $y$ & Prompt and generated completion \\
$L$, $t_i$, $\mathbf t(y)$ & Structure length, token $i$, full structure path \\
$d(\cdot,\cdot)$ & Normalized LFQ bit-Hamming distance \\
$z_i$, $b_{ik}$, $m_i$ & Token ID, its $k$-th bit, XOR mutation mask \\
\midrule
$\mathcal A$, $n_A$ & Anchor set and its size \\
$\mathcal M$, $n_M$ & Mutant pool and its size \\
$G_g$, $n_G$ & Structural group $g$ and number of groups \\
$S_g$, $n_R$ & Selected representatives in group $g$ and their count \\
\midrule
$R(x,y)$, $R_r$, $\omega_r$ & Total reward, reward component $r$, its weight \\
$b_g$, $s_M$ & Group baseline and mutant-pool scale \\
$A_i^M$, $A_a$ & Mutant and anchor advantages \\
$\mu_g$, $s_\mu$, $p_g$ & Group mean, between-group std, softmax weight \\
$W_g$, $w_{\min}$ & Group weight and its floor \\
\midrule
$\delta_i$, $\kappa$, $q_i$ & Distance proxy, calibration, pseudo-ratio \\
$\omega_i$, $\omega_i^-$ & Mutant weight and negative-advantage attenuation \\
\midrule
$K$, $E$, $Q$, $P_g$ & Population size, evaluated children, cheap pool, population \\
$\pi_\theta$, $\pi_{\mathrm{ref}}$, $J(\theta)$ & Policy, reference policy, KL-regularized objective \\
$\mathcal D$ & Prompt distribution over single-chain sequence targets \\
$\rho_a$, $\omega_a^{\mathrm{TOPR}}$ & Rollout ratio and TOPR weight \\
$\bar\ell_\theta$, $\mathcal L_{\mathrm{MGPO}}$, $\beta$ & Mean log-prob, MGPO loss, KL coefficient \\
\bottomrule
\end{tabular}
\end{table}

\section{Baseline methods}
\label{app:baselines}

All endpoint comparisons use the same SFT initialization and the same 183
held-out targets.  Unless stated otherwise, evaluation is policy-only with
$K=8$ samples per target and uses the same frozen DPLM-2 structural decoder.
The baselines are organized to separate ordinary on-policy RL, mutation with a
global reference, diversity-aware alternatives, and search-output imitation.

\paragraph{SFT initialization.}
This is the RL cold-start model before reinforcement learning and serves as the
zero-step reference for all reported RL deltas.

\paragraph{GRPO-8.}
GRPO-8 uses eight on-policy samples per prompt and applies the standard
group-relative on-policy update without evolutionary mutants.  It measures the
gain obtainable from ordinary small-group RL.

\paragraph{GRPO-64.}
GRPO-64 uses 64 on-policy samples per prompt.  Its 64 structurally evaluated
policy samples match the total number of expensive structural evaluations in
the default MGPO data-generation step (8 anchors plus 56 mutants), providing a
larger-budget on-policy control.

\paragraph{Mutant-GRPO.}
Mutant-GRPO trains on the 8-anchor plus 56-mutant pool but computes mutant
advantages relative to one pooled global mutant baseline.  It therefore tests
whether evolutionary mutation alone is sufficient when heterogeneous mutant
families share a single reward reference.

\paragraph{Same-layout global.}
Same-layout global is the closest causal control for MGPO.  It preserves the
anchor count, 56-mutant pool, reward-blind grouping, representative-selection
pipeline, group weighting, distance attenuation, and 32-sequence update layout,
but replaces every group baseline $b_g$ with one global mutant baseline
$b_{\rm global}$.

\paragraph{MGPO-local-only.}
This variant retains reward-blind structural grouping and the same
representative-selection pipeline, but removes the auxiliary group and distance
weights by setting $W_g=1$ and $\omega_i=1$.  Its mutant learning signal is
therefore determined by the structure-local baseline itself.

\paragraph{Mutant-GRPO+DivReward.}
This diversity-aware control augments the mutant-GRPO scalar reward with a
per-sample diversity contribution; the diversity coefficient is selected on
validation.  It tests whether explicit diversity reward shaping can replace the
change in credit-assignment reference used by MGPO.

\paragraph{ProteinZero-style.}
This is our adaptation of the pairwise diversity-regularization strategy used in
ProteinZero~\citep{wang2025proteinzero}.  It keeps the same initialization,
prompt stream, training horizon, and policy-only endpoint protocol while adding
an explicit pairwise diversity regularizer rather than changing the mutant
baseline.

\paragraph{SGRPO-style.}
This is our adaptation of the set-diversity objective in
SGRPO~\citep{ye2026sgrpo}.  The baseline explicitly rewards diversity at the
sample-set level and is evaluated under the same endpoint protocol as MGPO.

\paragraph{Mutant-AWR.}
Mutant-AWR uses the same anchors, 56-mutant pool, structural groups, and selected
24 mutant representatives as MGPO, but replaces signed local advantages with
positive normalized reward weights.  Its mutant objective is given in
Eq.~\eqref{eq:mutant-awr}, with $\tau$ selected from
$\{0.1,0.2,0.5,1.0\}$ on validation.  This control tests whether MGPO can be
explained simply as reward-weighted imitation of search outputs.

\paragraph{Full MGPO.}
The full method combines structure-local mutant advantages with group weighting
and negative-advantage distance attenuation as defined in
Eqs.~\eqref{eq:mutant-advantage}--\eqref{eq:mutant-offpolicy-weight}.  It is
included here only to make the design differences to each baseline explicit.

\section{Experimental protocol and audit details}
\label{app:experimental-protocol}

\subsection{Data split}
\label{app:data-audit}
Held-out targets are separated from the training data at the sequence-cluster
level: near-duplicate members of an existing sequence cluster are kept within a
single split.

\subsection{Component ablations}
\label{app:component-ablations}
Table~\ref{tab:component-ablations} reports the one-at-a-time component
ablations.  The distance heuristic and representative compression have the
smallest point-estimate effects among the ablated components.

\begin{table}[H]
\caption{One-at-a-time MGPO component ablations.}
\label{tab:component-ablations}
\centering\small
\setlength{\tabcolsep}{5pt}
\begin{tabular}{lrrrr}
\toprule
Variant & lDDT avg@8 $\uparrow$ & TM avg@8 $\uparrow$ & LFQ $\bar d$ $\uparrow$ & 3D div. $\uparrow$ \\
\midrule
Full MGPO & 0.7427 & 0.7532 & 0.0486 & 0.1518 \\
$-$ group-local advantage & 0.7145 & 0.7270 & 0.0352 & 0.1205 \\
$-$ reward-blind grouping & 0.7312 & 0.7410 & 0.0396 & 0.1330 \\
$-$ nonzero-mass floor & 0.7350 & 0.7440 & 0.0412 & 0.1372 \\
$-$ stratified selection & 0.7385 & 0.7480 & 0.0448 & 0.1420 \\
$-$ distance heuristic & 0.7400 & 0.7510 & 0.0472 & 0.1478 \\
$-$ 56-to-24 compression & 0.7410 & 0.7520 & 0.0479 & 0.1495 \\
\bottomrule
\end{tabular}
\end{table}

\subsection{Fixed-batch intervention: implementation details}
\label{app:fixed-batch}
The protocol is specified in Section~\ref{sec:exp-mechanism}; this appendix
records the implementation-level details.  Checkpoints are chosen before
inspecting the sign-flip outcome and sampled across early, middle, and late
training.  For every checkpoint and target batch we save the complete $8+56$
scored pool and the selected 24 representatives.  Two copies of the same model
and optimizer state receive one update with identical random-number states; the
only changed tensor is the mutant baseline.  Before the optimizer step we save
the flattened mutant-branch gradients to compute cosine similarity and norm
ratio.  After the step we re-score log probabilities of the same samples.

\section{Geometry, proxy, and backend audits}
\label{app:geometry-audit}

\paragraph{Within- versus between-group geometry.}
Reward-blind LFQ groups should be interpreted as algorithmic neighborhoods,
not automatically as biological states.  On saved mutant pools, the mean
within/between-group decoded distances are $0.112/0.284$ on validation and
$0.118/0.271$ on CAMEO22, with LFQ--3D adjusted Rand indices of $0.610$ and
$0.580$.  The substantially larger between-group than within-group decoded
distances, together with the moderate LFQ--3D adjusted Rand indices, indicate
that the reward-blind LFQ partition captures meaningful 3D geometry while not
being equivalent to a decoded-structure clustering.  We therefore interpret the
groups as algorithmic structural neighborhoods rather than as recovered
biological conformational states.

\paragraph{Proxy validity.}
The LFQ shift proxy has only moderate association with old-policy rarity:
Spearman $\rho(\delta,\Delta\mathrm{NLL})=0.46$ on validation and $0.43$ on
CAMEO22 pools.  Removing the proxy changes lDDT avg@8 from $0.7427$ to $0.7400$
in the ablation.  This is why the paper treats distance attenuation as a
stabilization heuristic rather than as a theoretical correction.

\subsection{Backend diagnostics}
\label{app:backend-diag}
\begin{figure}[H]
  \centering
  \includegraphics[width=\linewidth]{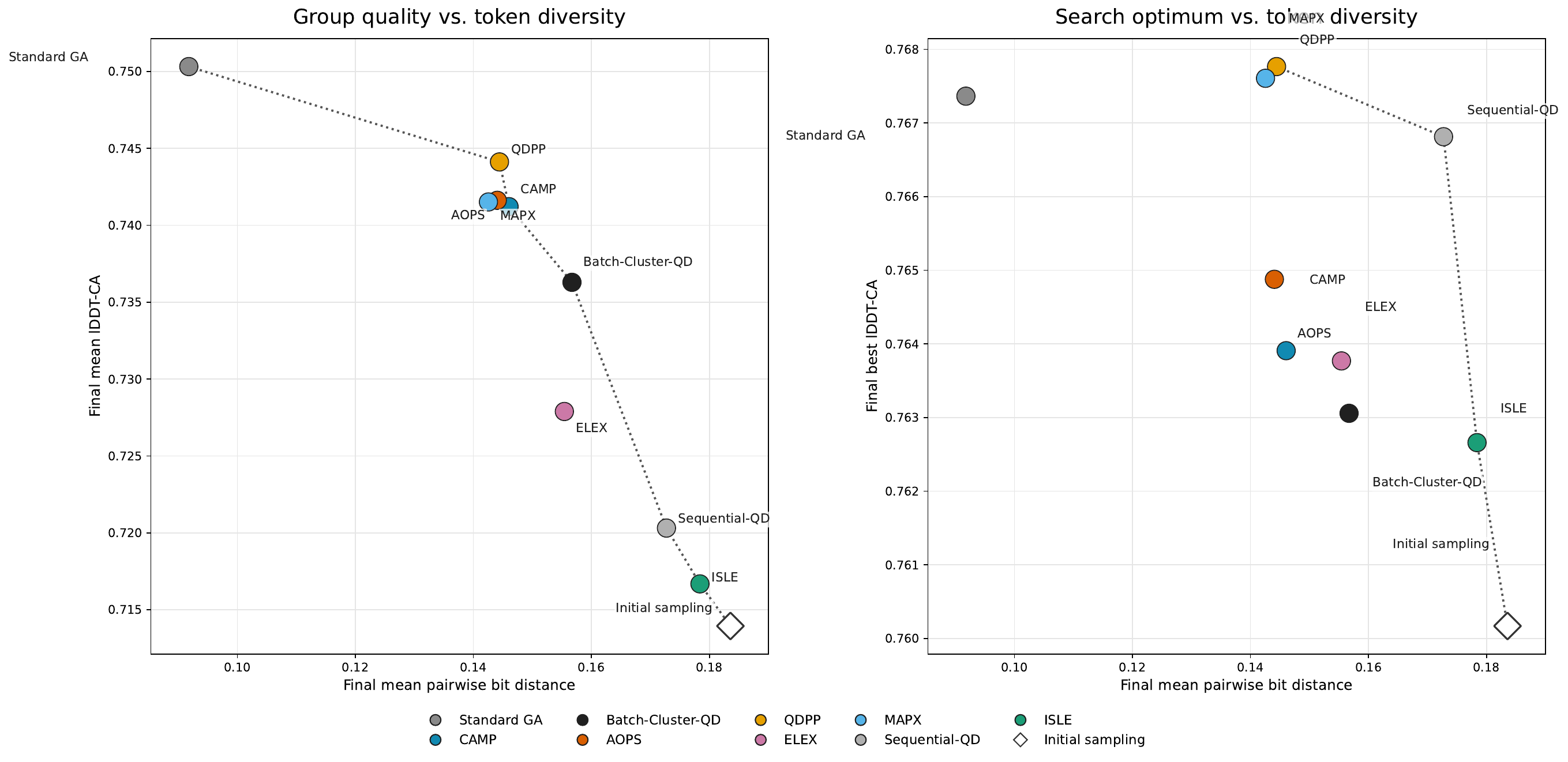}
  \caption{Low-budget mutation-backend ablation over the CAMEO22 benchmark.
  Each point is a backend-level aggregate after eight decoded child evaluations
  per target; the initial anchor is the diamond.  Left: final population mean
  lDDT-CA; right: final population best lDDT-CA.  Dotted lines connect the
  non-dominated frontier.}
  \label{fig:pareto}
\end{figure}

Before end-to-end RL, we characterize the nine candidate backends of
Table~\ref{tab:nine-strategies} at the search level.  These are backend
diagnostics, not comparisons of nine independent training objectives: the
downstream MGPO grouping and loss are held fixed.  All aggregates use
single-chain targets, population size $K=8$, and the same lDDT-CA evaluator.
Figure~\ref{fig:pareto} plots each backend's final mean and final best lDDT-CA
against mean pairwise token diversity.

%\begin{table}[H]
%\caption{Backend-level aggregate after eight decoded child evaluations per
%target.  lDDT-CA avg@8 andTM avg@8 are final population mean and best lDDT-CA;
%``LFQ div.'' is the scale-corrected final LFQ diversity; the initial anchor
%population has LFQ diversity $0.0400$.  Dashes mark entries not reported in
%the main scan.}
%\label{tab:backend-aggregate}
%\centering
%\small
%\begin{tabular}{lrrrr}
%\toprule
%Backend & lDDT-CA avg@8 & TM avg@8 & LFQ div. & 3D div. \\
%\midrule
%Standard GA      & 0.7503 & 0.7601 & 0.0412 & 0.1268 \\
%Sequential QD    & 0.7412 & 0.7518 & 0.0469 & 0.1447 \\
%Batch-Cluster QD & 0.7496 & 0.7592 & 0.0431 & 0.1334 \\
%CAMP             & 0.7451 & 0.7550 & 0.0446 & 0.1372 \\
%QDPP             & 0.7467 & 0.7570 & 0.0420 & 0.1321 \\
%ISLE             & 0.7398 & 0.7504 & 0.0473 & 0.1458 \\
%AOPS             & 0.7432 & 0.7532 & 0.0449 & 0.1381 \\
%ELEX             & 0.7474 & 0.7575 & 0.0452 & 0.1389 \\
%MAPX             & 0.7444 & 0.7544 & 0.0458 & 0.1402 \\
%\bottomrule
%\end{tabular}
%\end{table}

Standard GA has the highest final mean in this scan,
but its LFQ diversity is only marginally above the initial anchor value
 and its decoded 3D diversity is below the SFT reference in
Table~\ref{tab:main-results}. QDPP reaches the highest
final best score while retaining LFQ diversity on the corrected scale, illustrating why best-of-group and population-average quality should not be reported without a diversity axis. Sequential QD and ISLE retain still more diversity, but at a larger mean-quality cost.

\paragraph{Interpretation boundary.}
The native-reference reward and the diversity metrics above establish only
that the generated population is structurally broader under the stated quality
objective.  They do not establish recovery of biologically valid alternative
apo/holo or metastable states.  Accordingly, we restrict our claim to
structural population diversity under the stated quality objective and do not
interpret this evidence as recovery of biologically valid alternative
conformational states.

\section{Evolutionary backend strategies}
\label{app:backends}

This section provides full specifications for the nine mutation and survivor
strategies summarized in Table~\ref{tab:nine-strategies}.  Eight of them
adapt established algorithms from evolutionary
computation~\citep{holland1975adaptation,lehman2011novelty,pugh2016quality,cully2018quality,kaufman1990finding,kulesza2012dpp,whitley1999island,fialho2009aos,lacava2016epsilon,mouret2015mapelites},
while CAMP is a method-specific combination introduced in this work.

\subsection{Standard GA}
\label{app:standard-ga}

The control follows the canonical genetic-algorithm pattern of variation and
fitness-based survival~\citep{holland1975adaptation}.  Each generation mixes
three reward-blind proposal families: a CRF path assembled from group-level
transition statistics, an LFQ consensus/Fr\'{e}chet proposal, and unique
one-bit jumps from distinct parents.  After scoring, parents and children are
sorted by $R$, with deterministic generation order breaking ties, and the top
$K$ survive.  Thus Standard GA applies no explicit pressure for novelty; it
measures the quality gain and diversity collapse produced by pure elitism.

\subsection{Sequential QD}
\label{app:seqqd}

Sequential QD combines novelty search~\citep{lehman2011novelty} with the
quality-diversity principle of optimizing within behavioral
niches~\citep{pugh2016quality,cully2018quality}.  For archive $\mathcal H$,
we define the novelty of $z$ as the mean distance to its $k_n$ nearest
compatible archive members,
\begin{equation}
 n(z;\mathcal H)=\frac{1}{k'}\sum_{h\in
 \operatorname{kNN}_{k_n}(z,\mathcal H)}d(z,h),\qquad
 k'=\min(k_n,|\mathcal H|),
 \label{eq:novelty}
\end{equation}
with $k_n=3$.  Children are processed in deterministic proposal order.  A
child $c$ competes against its nearest current resident $r(c)$; the sole
population best is protected by redirecting competition to the second-nearest
resident.  The child replaces $r(c)$ if it is unique and either
\begin{equation}
 R(c)>R(r(c)),
 \quad\text{or}\quad
 n(c)\ge n(r(c))+\delta_n
 \ \wedge\
 R(c)\ge \max(R_{\min}^{0},R_{\max}^{g}-\delta_q).
 \label{eq:seqqd-accept}
\end{equation}
Every sufficiently novel probe may enter a bounded historical archive even
when it loses local competition.  When the archive exceeds its capacity, the
most crowded entry is removed.

\subsection{Batch-Cluster QD}
\label{app:batchqd}

Batch-Cluster QD asks whether broad mutation can be made evaluation-efficient.
It first generates $Q=64$ unique paths using crossover, LFQ-guided mutations,
consensus mutations, and occasional wider jumps.  Without observing child
reward, it reduces this pool to $E=8$ actual paths by farthest-first
$k$-medoids with two refinement passes~\citep{kaufman1990finding}.  Because a
medoid is an actual valid path, no synthetic cluster center must be decoded.
Only these medoids receive structural reward evaluations.
In the default MGPO configuration, this generation is repeated seven times per
prompt, so the seven $E=8$ evaluated medoid sets accumulate to the $n_M=56$
mutant pool used in Section~\ref{sec:mgpo}.  The $Q=64$ cheap pool is
regenerated per generation and does not consume structural evaluations.

Survivor selection uses $q$ quality slots (six when $K=8$), filled by the
highest-reward unique paths.  Each remaining slot greedily selects
\begin{equation}
 z^*=\arg\max_{z:R(z)\ge\phi_g}\min_{s\in S}d(z,s),
 \label{eq:maxmin}
\end{equation}
where $S$ is the partial survivor set.  The floor
$\phi_g=\max(\bar R^0-\delta_0,R_{\max}^g-\delta_g)$ prevents diversity slots
from accumulating arbitrarily poor structures; $\delta_g$ is annealed during
search.  This quality/crowding separation is related to diversity-preserving
multi-objective selection~\citep{deb2002nsgaii}, but uses explicit slots rather
than Pareto ranks.

\subsection{CAMP}
\label{app:camp}

CAMP changes where and how mutations are made.  At residue position $j$, let
\begin{equation}
 u_j=1-\max_v\frac{1}{K}\sum_{x\in P_g}\mathbb{I}[t_j(x)=v]
 \label{eq:camp-disagreement}
\end{equation}
be token disagreement.  We greedily select four positions with largest $u_j$
subject to a separation of at least $\max(1,\lfloor L/10\rfloor)$ residues.
Each center produces one local and one broad child.  Local mutation chooses a
parent and flips one or two entropy-weighted bits inside radius
$\max(1,\lfloor L/32\rfloor)$; broad mutation first performs uniform bitwise
crossover, then flips three to five bits inside radius
$\max(1,\lfloor L/10\rfloor)$.  For position $j$ and bit $b$, the sampling
weight is $p_{jb}(1-p_{jb})+\epsilon$, so variation targets uncertain group
coordinates without reading reward.  Survivors use the quality/max--min rule
in \eqref{eq:maxmin}.  CAMP is a method-specific combination of
multi-center disagreement sampling and QD selection.

\subsection{QDPP}
\label{app:qdpp}

Determinantal point processes (DPPs) assign high mass to subsets that are
jointly high-quality and mutually dissimilar~\citep{kulesza2012dpp}.  QDPP
generates the same $Q=64$ cheap pool as Batch-Cluster QD, but assigns proposal
$i$ the proxy
\begin{equation}
 \widetilde R_i=R(p_i)+\lambda_n d(\mathbf t_i,\mathbf t_{p_i}),
 \label{eq:qdpp-proxy}
\end{equation}
where $p_i$ is its nearest parent and $\lambda_n=0.03$.  The proxy uses only
already-known parent fitness; child structural reward is not queried.  Scores
are mapped to positive qualities $q_i$, and the similarity and DPP kernels are
\begin{equation}
 S_{ij}=\exp\!\left(-\frac{d(\mathbf{t}_i,\mathbf{t}_j)^2}{2\sigma^2}\right),
 \qquad L_{ij}=q_iS_{ij}q_j,
 \label{eq:dpp}
\end{equation}
where $\sigma$ is the median nonzero pairwise distance.  Greedy MAP inference
selects the $E$ proposals maximizing successive $\log\det L_S$, with the
highest-proxy item forced into the set.  After true evaluation, the same
greedy DPP rule is applied to parents and children using actual $R$, restricted
to an annealed quality band.

\subsection{ISLE}
\label{app:isle}

Island models preserve multiple evolutionary trajectories by restricting
interaction between subpopulations~\citep{whitley1999island}.  ISLE assigns
the $K=8$ lineages to four two-member islands according to their root-parent
index.  A child for island $r$ is formed by uniform crossover between its
resident lineage and the next island's lineage, followed by a population-guided
bit mutation.  This neighbor mating is the only information exchange; ISLE
does not use unrestricted global mating.  Within each island, survivor
selection first retains the highest-reward member and then the member farthest
from that island best, breaking distance ties by reward.  Any unused global
slots are filled by reward.

\subsection{AOPS}
\label{app:aops}

AOPS adapts operator selection online, following bandit-based adaptive
operator selection~\citep{fialho2009aos}.  Its arms are crossover, LFQ,
CRF-guided mutation, wide mutation, and targeted mutation.  Untried arms are
scheduled first; subsequently arm $a$ is selected by
\begin{equation}
 \operatorname{UCB}_a=\bar U_a+c_{\mathrm{ucb}}
 \sqrt{\frac{\log N}{N_a}},\qquad c_{\mathrm{ucb}}=0.20,
 \label{eq:aops-ucb}
\end{equation}
where $N_a$ is its past trial count.  After child evaluation, its utility is
\begin{equation}
 U(c)=R(c)-R(p_{\mathrm{near}})+\eta
 \min_{p\in P_g}d(c,p),\qquad \eta=0.10.
 \label{eq:aops-utility}
\end{equation}
Thus prior reward outcomes choose the next operator, but the chosen primitive
does not consult reward when selecting mutation bits.  Survivors follow the
same quality/max--min rule as CAMP.

\subsection{ELEX}
\label{app:elex}

Lexicase selection considers cases separately rather than collapsing them
into one objective, and $\epsilon$-lexicase admits near-best candidates in
continuous spaces~\citep{lacava2016epsilon}.  ELEX proposes population-guided
wide mutations.  For survivor selection, the cases are the reward components
$\{R_1,\ldots,R_J\}$ available to the trainer plus archive novelty from
\eqref{eq:novelty}; it never uses withheld answers or test labels.  For
each survivor draw, cases are randomly permuted.  Starting with all remaining
candidates, case $c$ retains
\begin{equation}
 \mathcal S\leftarrow\{i\in\mathcal S:
 f_{ic}\ge \max_{j\in\mathcal S}f_{jc}-\epsilon_c\},\qquad
 \epsilon_c=\max(\epsilon_{\mathrm{lex}},\operatorname{MAD}_{j\in\mathcal S}f_{jc}).
 \label{eq:elex}
\end{equation}
Filtering stops when one candidate remains or all cases are exhausted; one
survivor is sampled and removed, and the procedure repeats until $K$ are
selected.
The value of $\epsilon_{\mathrm{lex}}$ is reported in the implementation
details rather than fixed by the method definition.
ELEX consequently favors specialists that are strong on at least
one ordering of reward/diversity cases, rather than only high scalar reward.

\subsection{MAPX}
\label{app:mapx}

MAPX instantiates MAP-Elites~\citep{mouret2015mapelites}, a canonical
quality-diversity archive that stores the best solution in each behavioral
cell.  Its two descriptors are
\begin{equation}
 d_{\mathrm{init}}(\mathbf{t})=d(\mathbf{t},\mathbf{t}_{\mathrm{best}}^{0}),
 \qquad
 \Delta\rho(\mathbf{t})=\rho(\mathbf{t})-\rho_0,
 \quad
 \rho(\mathbf{t})=\frac{1}{L}\sum_i\frac{\operatorname{popcount}(t_i)}{13},
 \label{eq:mapx}
\end{equation}
where $\mathbf t_{\mathrm{best}}^0$ is the best anchor and $\rho_0$ is the
initial mean bit density.  Distance edges $(0.005,0.02,0.06)$ and density-shift
edges $(-0.015,0,0.015)$ define a $4\times4$ archive; malformed or
length-mismatched paths are quarantined in an overflow cell.  Each cell keeps
only its highest-reward member.  The next population takes cell elites in
descending reward order and, if fewer than $K$ cells are occupied, fills the
remainder from the best unselected candidates.  Mutation radius increases
periodically across proposals, encouraging transitions between neighboring
and more distant cells.

\section{Parallel Evolution}
\label{app:parallel-evolution}

The evolutionary backends of Appendix~\ref{app:backends} are specified at the
level of proposal and survivor mechanisms, but running them inside the MGPO
update adds two per-prompt costs on top of the anchor rollout: (i)
\emph{mutation construction}, which builds the cheap candidate pool, computes
the bit-Hamming distance matrix of \eqref{eq:hamming}, clusters with balanced
$k$-medoids, and draws mutants; and (ii) \emph{structural reward evaluation},
which decodes each candidate and runs the structure scorers.  This appendix
records how we parallelize the construction stage, the measured $10.4\times$
full-scale speedup this yields, and the profiling result that then locates the
dominant remaining cost in the evaluation stage rather than in construction.
It supplies the efficiency accounting summarized in
Section~\ref{sec:experiments}.
\subsection{Embarrassingly parallel construction}
\label{app:parallel-mutation}

Candidate construction is independent across prompts: the mutant pool of one
prompt is determined entirely by its anchor population and its per-prompt
seed, and no prompt reads another prompt's state during construction.  The
construction cost is concentrated in the LFQ bit-Hamming distance matrix and
the balanced $k$-medoids refinement of Appendix~\ref{app:grouping}; both are
implemented as C-level NumPy kernels that release Python's global interpreter
lock, so a thread pool can overlap them across prompts even though the
surrounding orchestration remains single-threaded Python.  This is the
data-parallel analogue of the island model's independent
subpopulations~\citep{whitley1999island}, with the parallelism placed over
prompts (each prompt is a separate search problem) rather than over lineages
within one search.

We therefore replace the sequential per-prompt loop with a fixed worker pool
that applies the unchanged candidate builder $f(\cdot)$ to every prompt in
parallel, returning the per-group candidate set $\mathcal C_g$.  Each task is
seeded by a deterministic function of the global step, the group index, and
the generation index, combined with a base seed; refill attempts add a fixed
prime offset to obtain a fresh stream.  Because this seed is deterministic and
$f$ is deterministic, the parallel build reproduces the sequential build up to
floating-point reduction order.  The per-prompt mutation output is therefore
unchanged, so the proposal and survivor mechanisms of
Appendix~\ref{app:backends} are untouched.  The worker count is the
configuration key \texttt{mgpo\_mutation\_workers}; a value of $0$ selects
$\min(16,\max(1,\texttt{os.cpu\_count()}))$ automatically, and the effective
pool size is capped at the number of live groups to avoid oversubscribing the
Ray controller.  Algorithm~\ref{alg:parallel-mutation} summarizes one parallel
generation.

\begin{algorithm}[t]
\caption{Parallel per-prompt mutation construction for one generation.}
\label{alg:parallel-mutation}
\begin{algorithmic}[1]
\State \textbf{Input:} live groups $\{(g,P_g)\}$, generation $t$, worker count $W$, base seed.
\State Compose tasks $\{(g,P_g,\xi_{g,t})\}$ with a deterministic per-group seed $\xi_{g,t}$.
\State Map $f(\cdot)$ over all tasks with a thread pool of $\min(W,|\text{tasks}|)$
       workers, collecting $\mathcal C_g=f(P_g;\xi_{g,t})$.
\State Mark a group failed if $|\mathcal C_g|<e$; otherwise append its candidates
       to the pending proposal batch.
\State Score the surviving proposals in one batched reward call (the evaluation
       stage, not construction).
\end{algorithmic}
\end{algorithm}

\subsection{Measured speedup at full scale}
\label{app:parallel-speedup}

Table~\ref{tab:parallel-fullscale} compares the pre-parallelization run with
the parallelized run at the full 4B configuration (8 GPUs, TP$=$1/DP$=$8,
training batch 256, the same MGPO $56$-mutant layout).  The intentionally
optimized stage is mutation construction, which drops from $427$\,s to $41$\,s,
a $10.4\times$ speedup; the GPU-side stage times are reported as observed
wall-clock rather than as paired utilization measurements.  Structural reward
evaluation is optimized independently and is summarized in
Table~\ref{tab:reward-acceleration}.
After construction parallelization alone, the end-to-end step shortens from
$1452$\,s to $1092$\,s ($-24.8\%$).  Applying the exact structural-reward
acceleration in Table~\ref{tab:reward-acceleration} reduces the exact-MGPO
step to approximately $845$\,s (the exact reward-evaluation stage is about
$175$\,s instead of the $160$\,s proxy value in
Table~\ref{tab:timing-breakdown}); the approximate proxy-reward variant
reaches approximately $830$\,s and is reported only as a systems upper bound.
In a separate profiling run, training throughput rises from $868$ to
$1102$ tok/s ($+27\%$) with parallelization; this throughput measurement
should not be interpreted as the exact reciprocal of the step-time reduction.
The remaining logged stages (advantage computation $0.19$\,s, weight update
$5.0$\,s, and rollout dump $12.0$\,s) are individually sub-$1\%$ and are
absorbed into the residual.
\begin{table}[h]
\caption{Full-scale before/after of the per-prompt mutation parallelization at
the 4B configuration (8 GPUs, training batch 256; seven pre-parallelization
steps versus five parallelized steps, means over steps).  ``Before'' is the
sequential construction run; ``After'' uses the thread pool of
Section~\ref{app:parallel-mutation}.  The GPU-side stages are reported for
completeness and are ``unchanged within variation'' rather than exact paired
measurements.  Training throughput is measured on a separate profiling run and
is reported as observed wall-clock rather than as a paired ratio to the step
time.}
\label{tab:parallel-fullscale}
\centering
\small
\setlength{\tabcolsep}{4pt}
\begin{tabular}{lrrr}
\toprule
Stage & Before (s) & After (s) & Change \\
\midrule
Mutation construction & 427 & 41 & $10.4\times$ faster \\
Structural reward evaluation & 469 & 469 & optimized separately (Table~\ref{tab:reward-acceleration}) \\
Rollout generation & 149 & 156 & $+5\%$ \\
Old log-probability & 154 & 163 & $+6\%$ \\
Actor update & 253 & 263 & $+4\%$ \\
\midrule
Total step & 1452 & 1092 & $-24.8\%$ \\
Training throughput (tok/s) & 868 & 1102 & $+27\%$ \\
\bottomrule
\end{tabular}
\end{table}

The $10.4\times$ construction speedup is the direct effect of the thread pool:
construction is the only stage whose wall-clock scales with the number of
prompts without GPU involvement, so overlapping its NumPy kernels across
prompts removes the largest single CPU serialization in the step.  The
unchanged reward-evaluation stage confirms that the speedup did not come from a
budget or evaluator change: the same $56$ mutants per prompt are still decoded
and scored in both runs.

\paragraph{Structural reward acceleration.}
Structural reward evaluation was then optimized independently of construction.
Table~\ref{tab:reward-acceleration} separates three correctness-preserving
changes from one approximate variant.  Shared decode caching lets lDDT-CA and
TM-align reuse one decode; TM-align subprocess parallelization increases the
worker pool from $8$ to $16$; and \texttt{torch.compile} accelerates the frozen
structure decoder after warm-up.  These changes preserve the full reward for
all $56$ mutants required by Eq.~\eqref{eq:mutant-advantage} and reduce the
structural scoring stage from $469$\,s to approximately $175$\,s.  The final
row, \emph{proxy-reward evolution}, is different: the $56$ mutants are first
ranked with a cheap lDDT-CA + token-accuracy proxy and full structural reward is
computed only for the selected mutant representatives (the 8 anchors are
already fully scored).  This mode reaches about $160$\,s but changes reward
accounting, so it is reported as an approximate engineering ablation rather
than as the implementation of the headline MGPO objective.

\begin{table}[h]
\caption{Structural reward evaluation acceleration at the full 4B
configuration (step-1 measurement; each row is cumulative).  The first four
rows preserve exact full-reward accounting for all 56 mutants; the final row is
an approximate proxy variant and is not used as evidence for the exact MGPO
objective unless separately validated.}
\label{tab:reward-acceleration}
\centering
\small
\setlength{\tabcolsep}{4pt}
\begin{tabular}{lcc}
\toprule
Optimization & Reward eval (s) & Saving \\
\midrule
Baseline (decode cache 32, TM-align 8 workers) & 469 & -- \\
$+$ shared decode cache & 305 & $-164$ \\
$+$ TM-align 16 workers & 235 & $-70$ \\
$+$ torch.compile (warm; exact reward) & $\sim$175 & $\sim$$-60$ \\
$+$ proxy-reward evolution (approx.) & $\sim$160 & $\sim$$-15$ \\
\bottomrule
\end{tabular}
\end{table}

\subsection{Resulting breakdown and next bottleneck}
\label{app:efficiency-breakdown}

After parallelization and structural reward acceleration, mutation construction
and decode/TM-align scoring are no longer leading costs.
Table~\ref{tab:timing-breakdown} uses the \emph{approximate proxy} path because
that was the fastest profiled configuration; it should therefore be read as a
systems upper-bound on throughput rather than as the timing of the exact
headline objective.  The exact compiled-decoder path differs by only about
15\,s in the reward-evaluation stage in this profile (approximately 175\,s
instead of 160\,s), the proxy path does not compute full $R_i$ for every mutant and is
therefore reported separately from the exact objective.
Individual stage times are
approximate and may not sum exactly to the measured total, which varies with
batch size, sequence length, and TM-align worker count.
\begin{table}[h]
\caption{Per-step wall-clock breakdown for the fastest profiled 4B
configuration, which includes the approximate proxy-reward path (8 GPUs,
TP$=$1 / DP$=$8, training batch 256, maximum response length 9216, with
parameter/gradient/optimizer offload).
Exact-MGPO scoring is approximately 15\,s slower in the reward-stage
profile.}
\label{tab:timing-breakdown}
\centering
\small
\setlength{\tabcolsep}{4pt}
\begin{tabular}{lrrl}
\toprule
Stage & Time (s) & Share & Device \\
\midrule
Structural reward evaluation (decode + TM-align) & $\sim$160 & 19\% & CPU \\
Actor update & 262 & 32\% & GPU \\
Old log-probability & 163 & 20\% & GPU \\
Rollout generation & 156 & 19\% & GPU \\
Mutation construction & 41 & 5\% & CPU \\
Other & $\sim$48 & 5\% & -- \\
\midrule
Total & $\sim$830 & 100\% & -- \\
\bottomrule
\end{tabular}
\end{table}

Two consequences follow.  First, after parallelization and the reward
acceleration, both mutation construction and structural reward evaluation are
no longer dominant, so further engineering should be judged by correctness
rather than by large throughput gains.  Second, the dominant costs are now the
standard GPU-side rollout and optimization stages, so the next efficiency
target is the actor/rollout pipeline rather than the CPU decode-and-score path;
the residual CPU decode/TM-align cost could still be reduced by an in-process
TM-score implementation.

\subsection{Interpretation and limitations}
\label{app:parallel-interpretation}

The parallel construction is a correctness-preserving engineering change whose
value is scheduling: it overlaps CPU-side mutation work across prompts while
the controller would otherwise idle, and it removes a sequential bottleneck
that grows with the number of prompts.  At a much smaller configuration (batch size 4, maximum response length
6784), the same change moves throughput by only $+2.3\%$, because
mutation construction is already a negligible fraction of the step.
The $10.4\times$ construction speedup becomes visible only once the number of
prompts is large enough that construction dominates the CPU.  The fractions in
Table~\ref{tab:timing-breakdown} are configuration-dependent (they shift with
batch size, sequence length, and whether TM-align is batched), and the
before/after numbers in Table~\ref{tab:parallel-fullscale} compare two runs on
the same model and batch size but with slightly different rollout scheduling
settings, so the GPU-side stages should be read as ``unchanged within
variation'' rather than as exact paired measurements.

\section{MGPO implementation details}
\label{app:mgpo-details}

This appendix provides the derivations and algorithmic details of the MGPO
components summarized in Section~\ref{sec:mgpo}.

\subsection{Problem setup details}
\label{app:setup-details}

The scalar reward used by training and survivor selection is the weighted sum
of available reward functions,
\begin{equation}
 R(x,y)=\sum_{r=1}^{J}\omega_r R_r(x,y),\qquad \omega_r\ge0,
 \label{eq:reward}
\end{equation}
with invalid or undecodable completions assigned zero mutation weight and
excluded from population statistics.  The four components are (i) \emph{token
accuracy}, the exact positional match rate between the predicted structure
tokens $\mathbf t(y)$ and the reference tokens obtained by encoding the native
structure with the frozen DPLM-2 encoder
($\frac{1}{L}\sum_{i}\mathbb I[t_i=t^{\mathrm{ref}}_i]$, with mismatched lengths
scored as zero); (ii) \emph{lDDT-CA}; (iii) \emph{TM-align} (TM-score via the
TMalign program), both computed by decoding $\mathbf t(y)$ back to coordinates
with the frozen decoder and comparing against the native backbone; and (iv) a
\emph{format} term that checks the structure block is present and
length-consistent.  Because the reference is the native structure of the prompt
itself, all methods share the same supervised signal, and the reported
diversity/quality differences cannot be attributed to a different reward.
For the exact MGPO objective, this full scalar $R(x,y)$ is computed for all
$n_M=56$ mutants before grouping and baseline computation.  A cheap proxy may
be used for proposal construction, but replacing $R$ by the proxy for members of
$\mathcal M$ defines an approximate variant and must be labeled separately.
The mutation interface operates directly on the LFQ binary code; the
structure-token representation is defined in
Section~\ref{sec:dplm2-tokenizer}.  The fixed per-prompt data flow of the
default MGPO update is summarized in Table~\ref{tab:mgpo-layout}.

\begin{table}[H]
\caption{Fixed per-prompt data flow of the default MGPO update.}
\label{tab:mgpo-layout}
\centering
\small
\begin{tabular}{lrrl}
\toprule
Data block & Evaluated & Trained & Role \\
\midrule
Policy anchors & 8 & 8 & On-policy TOPR group \\
Evolutionary mutants & 56 & 0 & Grouping and representative pool \\
Selected mutant representatives & -- & 24 & Three groups, eight per group \\
Final training batch & -- & 32 & Joint MGPO sequence loss \\
\bottomrule
\end{tabular}
\end{table}

\subsection{Reward-blind balanced structural grouping}
\label{app:grouping}

We construct the full $n_M\times n_M$ distance matrix with
\eqref{eq:hamming} and apply deterministic balanced $k$-medoids with
$k=n_G$~\citep{kaufman1990finding}:
\begin{equation}
 \mathcal M=G_1\cup\cdots\cup G_{n_G},\qquad
 G_g\cap G_h=\varnothing\ (g\ne h),\qquad |G_g|\ge n_R.
 \label{eq:mgpo-groups}
\end{equation}
The first medoid minimizes total distance to the pool and the remaining
$n_G-1$ medoids are initialized farthest-first.  Samples are assigned to the
nearest medoid.  If a cluster has fewer than $n_R$ members, a non-medoid point
is moved from a cluster above capacity by minimizing the induced
assignment-cost increase.  Each medoid is then updated to the real member with
minimum within-cluster distance, for at most $R_{\mathrm{ref}}$ refinement
rounds.  The clustering stage never reads reward.

\subsection{Reward-stratified coverage selection}
\label{app:selection}

Each $G_g$ is compressed to $n_R$ distinct training representatives $S_g$.
We rank its members by reward and create $n_S$ quantile strata, with a quota
of $n_q$ representatives from each stratum.  The medoid and highest-reward
member are included first when their stratum quotas permit.  Remaining slots
are filled greedily by
\begin{equation}
 \operatorname{Score}(i)=
 \lambda_{\mathrm{cov}}\,\Delta\operatorname{Cov}(i)
 +\lambda_{\mathrm{div}}\,\operatorname{Div}(i)
 +\lambda_{\mathrm{qual}}\,\operatorname{Qual}(i),
 \label{eq:representative-score}
\end{equation}
with $\lambda_{\mathrm{cov}}+\lambda_{\mathrm{div}}+\lambda_{\mathrm{qual}}=1$,
subject to the stratum quotas, where
\[
 \operatorname{Div}(i)=\min_{j\in S}d(i,j),\qquad
 \operatorname{Qual}(i)=\frac{R_i-R_{\min,g}}
 {\max(R_{\max,g}-R_{\min,g},\varepsilon)},
\]
and the marginal coverage improvement is
\begin{equation}
 \Delta\operatorname{Cov}(i)=\frac{1}{|G_g|}\sum_{u\in G_g}
 \left[d(u,S)-\min\{d(u,S),d(u,i)\}\right].
 \label{eq:coverage}
\end{equation}
This selection deliberately retains negative, near-baseline, and positive
local contrasts; MGPO is consequently not an elite-only behavior-cloning
method.  Because the full-cluster baseline is computed before this compression,
the selected subset need not itself have zero-mean advantage.

\subsection{Group baseline and anchor alignment}
\label{app:baseline}

The baseline of group $g$ uses every member of the full cluster, whereas the
common scale uses the complete mutant pool:
\[
 b_g=\frac{1}{|G_g|}\sum_{i\in G_g}R_i,\qquad
 s_M=\operatorname{Std}_{i\in\mathcal M}(R_i),
\]
\begin{equation}
 A_i^M=\frac{R_i-b_g}{s_M},\qquad i\in S_g.
 \label{eq:mutant-advantage-app}
\end{equation}
Computing $b_g$ before the $n_M\rightarrow n_G n_R$ compression prevents the
reward-stratified selector from shifting the advantage baseline.  Anchors
keep the native prompt-group GRPO advantage~\citep{shao2024deepseekmath},
denoted $A_a$, and form group $G_0=\mathcal A$.

Let $\mu_0$ be the anchor mean and $\mu_g=b_g$ for mutant groups.  With
$\bar\mu=(n_G+1)^{-1}\sum_{g=0}^{n_G}\mu_g$ and
$s_\mu=[(n_G+1)^{-1}\sum_g(\mu_g-\bar\mu)^2]^{1/2}$, we compute the
softmax stably after subtracting the maximum group mean:
\[
 z_g=\frac{\mu_g-\max_h\mu_h}{s_\mu},\qquad
 p_g=\frac{\exp(z_g)}{\sum_h\exp(z_h)}.
\]
If $s_\mu=0$, we set $p_g=1/(n_G+1)$ and hence $W_g=1$ for every
group.  Otherwise, $W_g$ follows \eqref{eq:group-weight}.  Subtracting the
maximum prevents overflow, and the $s_\mu=0$ branch removes the
undefined near-zero-temperature case.

\subsection{Distance proxy calibration}
\label{app:proxy-calibration}

Let $\widetilde\delta$ be the median positive distance in the mutant pool.
We calibrate
\begin{equation}
 \kappa=-\frac{\log q_{\mathrm{med}}}{\widetilde\delta},\qquad
 q_{\mathrm{med}}\in(0,1),\qquad q_i=\exp(-\kappa\delta_i),
 \label{eq:pseudo-ratio}
\end{equation}
with $q_i=1$ when all distances vanish.  Negative-advantage attenuation
follows \eqref{eq:mutant-offpolicy-weight}.  We restrict the attenuation
parameters to $0<b\le a$; then for every $q\in(0,1]$,
$\log(q^a e^{b(1-q)})\le(b-a)(1-q)\le0$, so the multiplicative weight lies
in $(0,1]$.  The default is $a=b=1$.  This bound establishes only that the
heuristic attenuates negative updates; it does not make the mutant branch an
unbiased off-policy estimator.

\subsection{Training procedure details}
\label{app:training-details}

The optimization step is described in Section~\ref{sec:mgpo}.  Candidate paths must be
distinct from anchors, earlier mutants, and their current generation.  A
deterministic refill with new seeds is attempted at most $R_{\mathrm{refill}}$
times; failure to obtain $e$ unique valid mutants aborts the update rather than
padding with duplicates.  Evolutionary early stopping is disabled so that the
complete $n_M$-mutant pool is available.  Cheap token-space proposals do not
consume structural evaluations.  Metadata are gathered across data-parallel
workers before grouping, and the fixed $n_A+n_G n_R$ sequence layout is
re-sharded uniformly.

\section{Scope and interpretation}
\label{sec:scope}

MGPO's core claim is that structure-defined local baselines can preserve
learning signals from multiple mutant families while retaining a fixed,
well-audited training layout.  Its anchor branch is rollout-based TOPR, but
its mutant branch is not an unbiased off-policy policy-gradient estimator.
LFQ distance may not perfectly track 3D structural distance or policy
likelihood, and $q_i$ in \eqref{eq:pseudo-ratio} is not an importance ratio.  A strict correction
would require a tractable recorded mutation density.  We therefore present
distance mixtures, group count, group-weight floor, representative-selection
terms, mutant-pool size, and the nine evolutionary backends as explicit
design choices rather than as separate headline algorithms.

\end{document}